\documentclass[11pt]{article}

\usepackage[preprint]{acl}

\usepackage{times}
\usepackage{latexsym}

\usepackage[T1]{fontenc}
\usepackage[utf8]{inputenc}

\usepackage{microtype}

\usepackage{inconsolata}
\usepackage{amsmath}
\usepackage{hyperref}

\usepackage{graphicx}

\title{The Script Tax: Measuring Tokenization-Driven Efficiency and Latency Disparities in Multilingual Language Models}

\author{Aradhya Dixit \\
  Wake Technical Community College \\
  \texttt{adixit1@my.waketech.edu} \\\And
  Shreem Dixit \\
  University of North Carolina Charlotte \\
  \texttt{sdixit6@charlotte.edu} \\}

\begin{document}
\maketitle

\begin{abstract}
Pretrained multilingual language models are often assumed to be script-agnostic, yet their tokenizers can impose systematic costs on certain writing systems. We quantify this \emph{script tax} by comparing two orthographic variants with identical linguistic content. Across mBERT and XLM-R, the higher-fragmentation orthography shows a $\sim$3.4$\times$ increase in fertility (6.73--6.85 vs.\ 2.10--2.35 tokens/word), leading to a 16.5$\times$ inference slowdown (0.23 vs.\ 3.8 sentences/second) on identical hardware. Using bits per character (BPC) to avoid the ``NLL paradox'' from subword fragmentation, we find a substantial increase in information cost: +19.7\% for mBERT (8.06$\rightarrow$9.65) and +47.1\% for XLM-R (12.19$\rightarrow$17.94). A round-trip conversion check (CER$_{rt}$=0.31) suggests these gaps reflect orthography-conditioned processing rather than mapping noise. Our results highlight tokenization as a key source of inequity in multilingual NLP and motivate script-aware tokenization and pretraining.
\end{abstract}

\section{Introduction}

Large multilingual language models are often treated as \emph{script-agnostic}: if two inputs express the same linguistic content, we might expect comparable model quality and comparable inference cost regardless of the writing system \citep{ahia-etal-2023-languages,petrov-etal-2023-tokenizers}. In practice, most multilingual models rely on a single pretrained subword tokenizer whose vocabulary and segmentation behavior can differ drastically across scripts \citep{limisiewicz-etal-2023-tokenization,petrov-etal-2023-tokenizers}. When a tokenizer fragments one script into many more pieces than another, the model is forced to process longer sequences for the same content, increasing memory use, compute cost, and latency---and potentially degrading modeling efficiency by obscuring word- and morpheme-level structure \citep{ahia-etal-2023-languages,limisiewicz-etal-2023-tokenization}.

We refer to this systematic penalty as a \emph{script tax}: a writing-system-dependent increase in (i) \textbf{tokenization cost}, (ii) \textbf{computational cost}, and (iii) \textbf{information cost} \citep{ahia-etal-2023-languages,petrov-etal-2023-tokenizers}. The core mechanism is \textbf{tokenization fragmentation}, which we quantify with \emph{fertility} (tokens per word). Higher fertility inflates the effective sequence length, which can induce disproportionate slowdowns in Transformer inference due to attention costs that scale superlinearly with sequence length \citep{ahia-etal-2023-languages,petrov-etal-2023-tokenizers}. Importantly, standard log-likelihood metrics can be misleading under severe fragmentation: predicting many small subword fragments can yield deceptively low token-level loss even when the model is less efficient at representing the underlying information \citep{limisiewicz-etal-2023-tokenization,petrov-etal-2023-tokenizers}. To address this, we evaluate modeling efficiency using \textbf{bits per character (BPC)}, normalizing the loss by the number of characters rather than the number of tokens.

In this work, we instantiate the script tax using paired sentence sets that preserve identical linguistic content across two orthographic variants and evaluate two widely used multilingual masked language models (mBERT and XLM-R). We observe a large fertility gap: 6.73--6.85 tokens/word for the higher-fragmentation orthography versus 2.10--2.35 for the lower-fragmentation orthography (a $\sim$3.4$\times$ increase). This tokenization disparity translates into a substantial runtime penalty on identical hardware, reducing throughput from $\sim$3.8 to $\sim$0.23 sentences/second (16.5$\times$ slower). After normalizing by characters, we find a pronounced information-cost increase: BPC rises by +19.7\% for mBERT (8.06$\rightarrow$9.65) and +47.1\% for XLM-R (12.19$\rightarrow$17.94). A round-trip conversion robustness check (CER$_{rt}$=0.31) suggests these differences are not artifacts of mapping instability, but reflect orthography-conditioned processing induced by pretrained tokenizers.

\begin{figure*}[t]
  \centering
  \includegraphics[width=\textwidth]{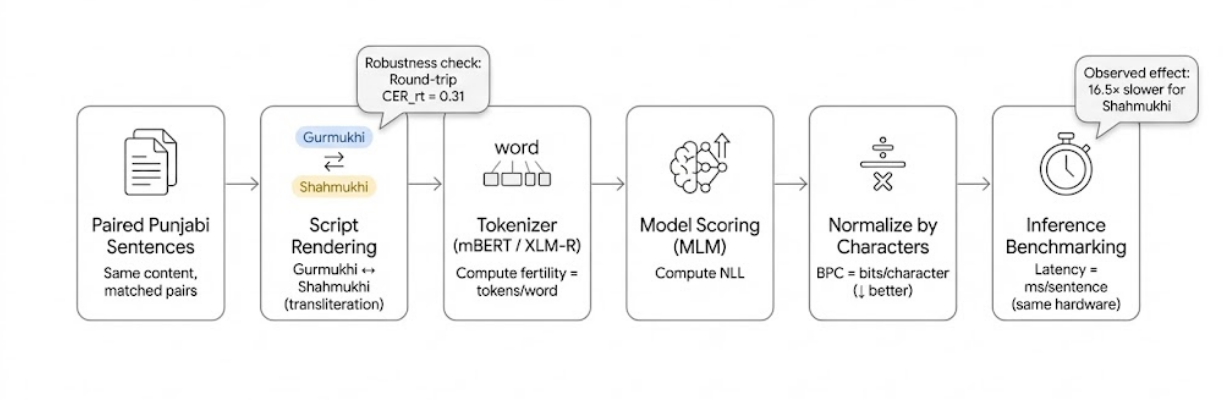}
  \caption{Evaluation pipeline used to measure the \emph{script tax}. We compare paired sentences across orthographic variants and compute (i) tokenization fertility (tokens/word), (ii) modeling efficiency via BPC (loss normalized by character count), and (iii) inference latency/throughput on identical hardware.}
  \label{fig:pipeline}
\end{figure*}

Our contributions are:
\begin{itemize}
  \item We define and operationalize the \textbf{script tax} as a joint disparity in fertility, latency, and character-normalized information cost (BPC) under controlled, content-matched evaluation.
  \item We show that \textbf{tokenization fragmentation} is a first-order driver of both compute overhead (16.5$\times$ slowdown) and reduced modeling efficiency (up to +47.1\% BPC), even when token-level NLL appears favorable.
  \item We provide a simple \textbf{robustness check} (round-trip CER) that helps separate script-induced effects from conversion noise.
\end{itemize}

\section{Methodology}
\label{sec:method}

Figure~\ref{fig:pipeline} summarizes our pipeline. We quantify the \emph{script tax} along three axes: tokenization fragmentation, inference overhead, and character-normalized information cost.

\subsection{Paired setup}
Let $\mathcal{D}=\{(x_A^{(n)},x_B^{(n)})\}_{n=1}^{N}$ be paired sentences with identical content in orthographies $A$ and $B$. For model $m$ with tokenizer $\tau_m(\cdot)$, define token length $L_m(x)=|\tau_m(x)|$ and word count $W(x)$ (whitespace tokens).

\subsection{Fragmentation (fertility)}
We measure fragmentation via fertility (tokens/word):
\begin{equation}
\begin{aligned}
F_m(x) &= \frac{L_m(x)}{W(x)}, \\
\Delta F_m &= \frac{1}{N}\sum_{n}\Big(F_m(x_B^{(n)})-F_m(x_A^{(n)})\Big).
\end{aligned}
\end{equation}

\subsection{Information cost (BPC)}
Token-level NLL can be biased by subword fragmentation, so we normalize by characters. Let $C(x)$ be the number of Unicode characters (excluding spaces). For a fixed masking scheme, let $\mathrm{NLL}_m(x)$ be the average masked-token loss in nats. We report bits per character:
\begin{equation}
\begin{aligned}
\mathrm{BPC}_m(x) &=
\frac{\mathrm{NLL}_m(x)}{\log 2}\cdot\frac{|M(x)|}{C(x)}, \\
\delta \mathrm{BPC}_m &=
\frac{\overline{\mathrm{BPC}}_m^{(B)}-\overline{\mathrm{BPC}}_m^{(A)}}{\overline{\mathrm{BPC}}_m^{(A)}}.
\end{aligned}
\end{equation}

\subsection{Compute cost (latency)}
Let $T_m(x)$ be measured inference time under fixed hardware and settings. We report median latency and the latency tax:
\begin{equation}
\mathrm{Lat}_m^{(s)}=\mathrm{median}_{n}\,T_m(x_s^{(n)}), \qquad
\rho_m^{\mathrm{lat}}=\frac{\mathrm{Lat}_m^{(B)}}{\mathrm{Lat}_m^{(A)}}.
\end{equation}
Sequence inflation increases attention cost roughly quadratically:
\begin{equation}
\mathrm{Cost}(x)\propto L_m(x)^2 \;\;\Rightarrow\;\;
\frac{\mathrm{Cost}(x_B)}{\mathrm{Cost}(x_A)}\approx\left(\frac{L_m(x_B)}{L_m(x_A)}\right)^2.
\end{equation}

\subsection{Robustness}
We validate conversion stability using round-trip CER. With maps $\pi_{A\rightarrow B}$ and $\pi_{B\rightarrow A}$, let $\hat{x}_A=\pi_{B\rightarrow A}(\pi_{A\rightarrow B}(x_A))$. Using edit distance $\mathrm{ED}$:
\begin{equation}
\mathrm{CER}_{rt}=\frac{1}{N}\sum_{n}\frac{\mathrm{ED}(x_A^{(n)},\hat{x}_A^{(n)})}{C(x_A^{(n)})}.
\end{equation}

We summarize the script tax per model as $(\Delta F_m,\rho_m^{\mathrm{lat}},\delta \mathrm{BPC}_m)$.

\section{Results}
\label{sec:results}

We report results for two multilingual masked language models (mBERT and XLM-R) under the paired evaluation described in Section~\ref{sec:method}. All measurements are computed on matched sentence pairs and identical inference settings.

\subsection{Tokenization bottleneck: fertility gap}
\begin{figure}[t]
  \centering
  \includegraphics[width=\columnwidth]{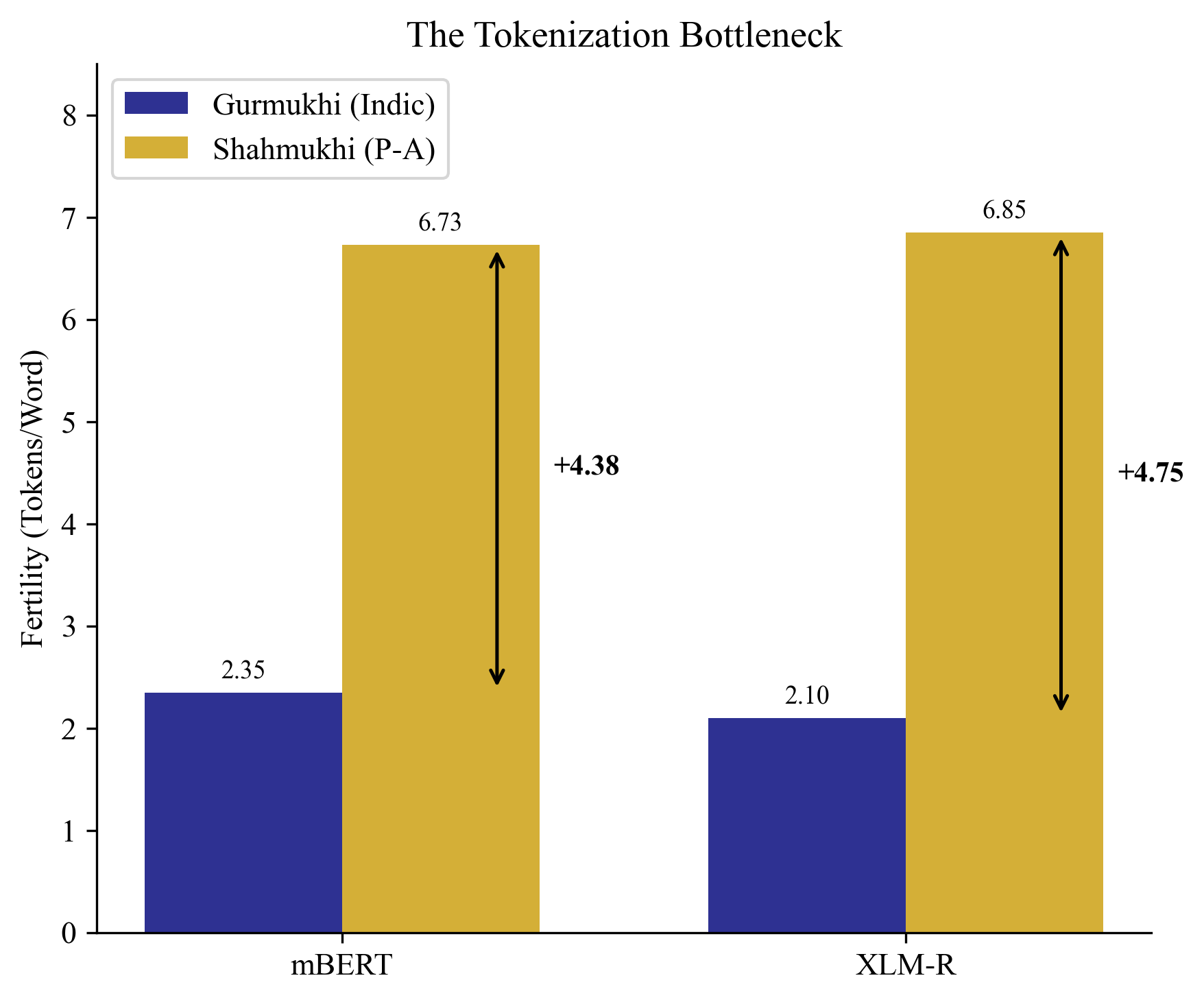}
  \caption{Tokenization bottleneck (fertility, tokens/word). The higher-fragmentation orthography requires substantially more tokens per word across both mBERT and XLM-R.}
  \label{fig:fertility}
\end{figure}

Figure~\ref{fig:fertility} shows a large and consistent fragmentation disparity across models. The higher-fragmentation orthography requires 6.73--6.85 tokens/word, while the lower-fragmentation orthography requires 2.10--2.35 tokens/word, yielding a gap of +4.38 to +4.75 tokens/word ($\sim$3.4$\times$ longer sequences). This indicates that the tokenizer is effectively operating closer to character-level segmentation for one orthography, which inflates sequence length and drives downstream compute overhead.

\subsection{Computational overhead: latency tax}
Sequence inflation yields a substantial runtime penalty. Measured on identical hardware, throughput drops from $\sim$3.8 to $\sim$0.23 sentences/second, corresponding to a 16.5$\times$ slowdown for the higher-fragmentation orthography. This is consistent with the superlinear dependence of Transformer inference cost on sequence length (Section~\ref{sec:method}), where longer token sequences amplify attention and memory overhead.

\subsection{Information cost: BPC ``script tax''}
\begin{figure}[t]
  \centering
  \includegraphics[width=\columnwidth]{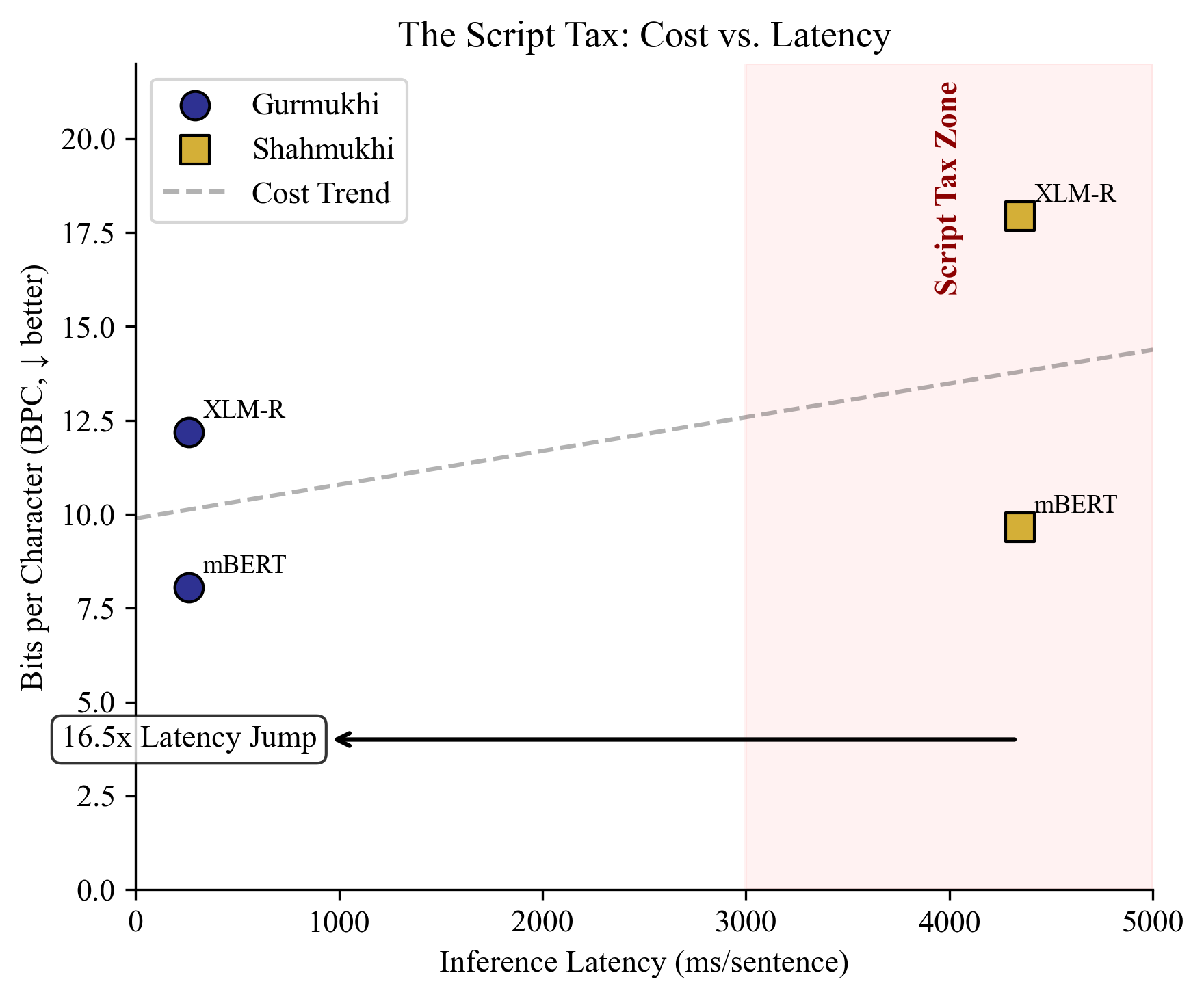}
  \caption{The script tax as a joint disparity in information cost and runtime. The higher-fragmentation orthography exhibits both higher BPC (worse; $\downarrow$ better) and substantially higher inference latency.}
  \label{fig:script-tax}
\end{figure}

Figure~\ref{fig:script-tax} plots the joint effect of information cost (BPC; lower is better) and inference latency. Character-normalized efficiency degrades substantially for the higher-fragmentation orthography: BPC increases by +19.7\% for mBERT (8.06$\rightarrow$9.65) and by +47.1\% for XLM-R (12.19$\rightarrow$17.94). This confirms that the apparent token-level ``ease'' sometimes observed under fragmentation is an artifact of predicting smaller units; when normalized to characters, the model is less efficient at representing the same underlying content.

\subsection{Summary}
Across both models, tokenization fragmentation is the dominant upstream driver: a $\sim$3.4$\times$ increase in sequence length produces a 16.5$\times$ slowdown and a large rise in character-normalized information cost (up to +47.1\% BPC). Together, these results demonstrate that pretrained tokenizers can introduce structural disparities in both computational accessibility and modeling efficiency.

\section{Discussion and Implications}
\label{sec:discussion}

Our results show that tokenization is not a neutral preprocessing choice in multilingual LMs: it can create a large, systematic \emph{cost and quality disparity} between orthographic variants with identical linguistic content \citep{ahia-etal-2023-languages,petrov-etal-2023-tokenizers,limisiewicz-etal-2023-tokenization}.

\subsection{One upstream cause, multiple downstream taxes}
The fertility gap (Figure~\ref{fig:fertility}) is the dominant upstream driver: because the higher-fragmentation orthography expands sequence length by $\sim$3.4$\times$, even modest superlinear scaling in the forward pass can translate into large slowdowns, which we observe as a 16.5$\times$ throughput drop. This gap is not a ``deployment detail'': it directly affects user experience (latency) and provider cost (compute).

\subsection{Why token-level loss can be misleading (and BPC fixes it)}
Methodologically, fragmentation can produce an \emph{NLL paradox}: predicting many short fragments can reduce per-token loss while still representing the same content less efficiently. BPC counters this by normalizing by characters, making the comparison invariant to how aggressively the tokenizer splits; the observed BPC increases (+19.7\% for mBERT and +47.1\% for XLM-R; Figure~\ref{fig:script-tax}) therefore indicate a genuine degradation in information efficiency, not a metric artifact.

\subsection{Robustness and interpretation}
The round-trip CER check (CER$_{rt}=0.31$) is not perfect reconstruction, but it provides evidence that the paired setup is not dominated by conversion noise; importantly, the same orthography that yields higher fertility also yields both higher latency and higher BPC, supporting the interpretation that the observed gaps are primarily orthography-conditioned processing effects rather than dataset artifacts.

\subsection{Practical implications}
Practically, these findings suggest two lessons \citep{ahia-etal-2024-magnet,liang-etal-2023-xlm,tao-etal-2024-vocabulary}: (i) \textbf{script-aware preprocessing/tokenization matters}---for underrepresented scripts, adapting the tokenizer (e.g., vocabulary augmentation or re-tokenization) may provide large gains in both inference cost and modeling efficiency without changing the model architecture \citep{ahia-etal-2024-magnet,liang-etal-2023-xlm,tao-etal-2024-vocabulary}; and (ii) \textbf{report compute-aware metrics}---evaluations that only report token-level perplexity/NLL risk hiding real inequities, whereas reporting fertility, character-normalized loss (BPC), and latency gives a more faithful picture of accessibility and performance \citep{ahia-etal-2023-languages,petrov-etal-2023-tokenizers,ramesh-etal-2023-fairness}. Overall, the script tax is best understood as a structural property of pretrained tokenizers: it creates a predictable mapping from sequence inflation $\rightarrow$ compute penalty $\rightarrow$ degraded character-normalized efficiency, raising both fairness and efficiency concerns for multilingual NLP.

\section{Conclusion}
\label{sec:conclusion}

We introduced the \emph{script tax}, a tokenization-driven disparity in both compute cost and modeling efficiency between orthographic variants with the same linguistic content. Across mBERT and XLM-R, we observe a large fertility gap (6.73--6.85 vs.\ 2.10--2.35 tokens/word) that induces a 16.5$\times$ inference slowdown on identical hardware, and a substantial increase in character-normalized information cost (BPC): +19.7\% for mBERT and +47.1\% for XLM-R. A round-trip CER check suggests these effects are not dominated by conversion noise.

Practically, our findings imply that reporting only token-level loss can obscure real disparities created by tokenizer fragmentation, and that character-normalized metrics such as BPC should be paired with compute measures (latency/throughput) when assessing multilingual model quality. More broadly, these results motivate script-aware interventions---such as tokenizer adaptation or script-balanced pretraining---as straightforward ways to reduce both the inference burden and the representational inefficiency imposed on higher-fragmentation scripts \citep{ahia-etal-2024-magnet,liang-etal-2023-xlm,tao-etal-2024-vocabulary,xue-etal-2022-byt5}.
\section{Limitations}
\label{sec:limitations}

Our analysis is limited to two orthographic variants and two multilingual masked language models (mBERT and XLM-R), so the magnitude of the script tax may differ for other scripts, tokenizers, and model families. The paired setup relies on an orthography conversion pipeline; while round-trip CER (CER$_{rt}$) offers a robustness check, residual conversion artifacts may still influence results. Finally, latency numbers depend on a specific hardware/inference configuration, though the relative penalties from sequence inflation should generalize qualitatively.

% Bibliography entries for the entire Anthology, followed by custom entries
%\bibliography{anthology,custom}
% Custom bibliography entries only
\nocite{*}

\bibliography{custom}

\end{document}